\documentclass[twoside,11pt]{article}
\usepackage[margin=1in]{geometry}
\usepackage[round]{natbib}
\usepackage[pdfusetitle]{hyperref}

\usepackage{latexsym,amsmath,amssymb,amsthm}
\usepackage{thmtools}

\usepackage{url}
\usepackage{tikz}
\usepackage{xspace} 
\usepackage{footnote}
\usepackage{framed}
\usepackage{ifthen}
\usepackage{cleveref}
\usepackage{caption}
\usepackage{subcaption}

\usetikzlibrary{arrows,automata,arrows.meta}
\usetikzlibrary{positioning}
\usetikzlibrary{backgrounds}
\usetikzlibrary{calc}
\usetikzlibrary{patterns}

\newcommand\p[1]{\ensuremath{\left( #1 \right)}} 

\newcommand{\y}{y}
\newcommand{\var}[1]{
  \def\temp{#1}
  \ifx\temp\empty\y
  \else\y_{{\temp}}
  \fi
}

\newcommand{\flow}[1]{\phi_{#1}}

\newcommand{\vars}{{\by}\xspace}
\def\Iverson#1{\left[ #1 \right]}
\newcommand{\ivar}[1]{\Iverson{\var{#1}}\xspace}
\newcommand{\coeff}[1]{
  \def\temp{#1}
  \ifx\temp\empty c
  \else c_{{\temp}}
  \fi
}
\newcommand{\coeffs}{{\bc}}
\newcommand{\connector}{:}
\newcommand{\varlab}[2]{\var{#1\connector#2}}
\newcommand{\coefflab}[2]{\coeff{#1\connector#2}}

\newcommand{\labelString}{\mbox{\footnotesize label}}

\newcounter{constraint}
\newenvironment{constraint}[1]
{\refstepcounter{constraint}%
  \begin{framed}
    \noindent{\bf Constraint~\theconstraint:} #1
    \begin{eqnarray*}
      }
      {
    \end{eqnarray*}
  \end{framed}
}

\newcommand{\lbl}[1]{{\tt #1}}
\newcommand{\false}{\lbl{false}\xspace}
\newcommand{\true}{\lbl{true}\xspace}
\newcommand{\bx}{{\bf x}}
\newcommand{\by}{{\bf y}}
\newcommand{\bc}{{\bf c}}
\newcommand{\sY}{\mathcal{Y}}
\newcommand{\sS}{\mathcal{S}}
\newcommand{\sL}{\mathcal{L}}

\def\Sum{\sum\limits}

\def\Vee{\bigvee\limits}
\def\Wedge{\bigwedge\limits}

\ifthenelse{\isundefined{\definition}}{\newtheorem{definition}{Definition}}{}
\ifthenelse{\isundefined{\assumption}}{\newtheorem{assumption}{Assumption}}{}
\ifthenelse{\isundefined{\proposition}}{\newtheorem{proposition}{Proposition}}{}
\ifthenelse{\isundefined{\theorem}}{\newtheorem{theorem}{Theorem}}{}
\ifthenelse{\isundefined{\lemma}}{\newtheorem{lemma}{Lemma}}{}
\ifthenelse{\isundefined{\corollary}}{\newtheorem{corollary}{Corollary}}{}
\ifthenelse{\isundefined{\alg}}{\newtheorem{alg}{Algorithm}}{}
\ifthenelse{\isundefined{\example}}{\declaretheorem[style=definition,qed=$\square$,sibling=definition]{example}}{}
\ifthenelse{\isundefined{\conjecture}}{\newtheorem{conjecture}{Conjecture}}{}
\ifthenelse{\isundefined{\claim}}{\newtheorem{claim}{Claim}}{}
\ifthenelse{\isundefined{\property}}{\newtheorem{property}{Property}}{}

\def\naive{na\"{\i}ve\xspace}

\def\naively{na\"{\i}vely\xspace}	

\begin{document}

\title{The Integer Linear Programming Inference Cookbook}

\author{
Vivek Srikumar \\
University of Utah
\and
Dan Roth\\
University of Pennsylvania
}

\date{}

\maketitle

\begin{abstract}
  Over the years, integer linear programs have been employed to
  model inference in many natural language processing problems. This
  survey is meant to guide the reader through the process of framing
  a new inference problem as an instance of an integer linear
  program and is structured as a collection of recipes. At the end,
  we will see two worked examples to illustrate the use of these
  recipes.
\end{abstract}

\tableofcontents

\newpage 

\section{Introduction}
\label{sec:introduction}

Effective decision-making requires the use of knowledge. This has been a clear,
and long-standing principle in AI research, as reflected, for example, in the
seminal early work on knowledge and AI---summarized
by~\citet{brachman1985readings}---and the thriving \emph{Knowledge
  Representation and Reasoning} and the \emph{Uncertainty in AI}
communities. However, the message has been somewhat diluted as data-driven
statistical learning has become increasingly pervasive across
AI\@. Nevertheless, the idea that reasoning and learning need to work
together~\citep{khardon1996reasoning,roth1996learning} and that knowledge representation is a
crucial bridge between them has not been lost.

One area where the link between learning, representation, and reasoning has been
shown to be essential and has been studied extensively is Natural
Language Processing (NLP), and in particular, the area of Structured Output
Prediction within NLP\@. In structured problems, there is a need to assign
values to multiple random variables that are interrelated. Examples include
extracting multiple relations among entities in a document, where a the two
arguments for a relation
such as \lbl{born-in} cannot refer to people, or co-reference
resolution, where gender agreement must be maintained when determining that a
specific pronoun refers to a given entity. In these, and many other such
problems, it is natural to represent knowledge as Boolean functions over
propositional variables. These functions would express knowledge, for example,
of the form ``if the relation between two entities is \lbl{born-in}, then its
arguments must be a \lbl{person} and a \lbl{location}'' (formalized as functions
such as $x_i \rightarrow x_j \lor x_k$, or exactly one of $x_1, x_2, \ldots x_k$
can be \true).  These functions serve to \emph{constrain} the feasible solutions
to the inference problem and open the possibility to model the global decision
problem as a constrained optimization problem.

An influential, and as we will see, also natural formalism for the decision
problem is to frame it as an Integer Linear Program (ILP). This approach was
first employed in NLP in the context of information extraction and machine
translation~\citep{roth2004linear,germann2004fast,roth2005integer} The objective
function for the integer program in question is typically learned, and could be
viewed as proposing, for each variable of interest, a distribution over the
values it can take. The final assignment to these variables is then determined
by maximizing the objective, subject to knowledge constraints, such as the ones
described above. The ability to decouple the modeling of a problem and the
knowledge needed to support inference, from learning the models is one reason
that has made the ILP formulation a popular one in NLP\@. Over the years, ILPs
have been employed to model inference in many natural language processing (NLP)
problems---information
extraction~\citep{roth2004linear,choi2006joint,denis2011predicting,berant2014modeling},
decoding in machine translation~\citep{germann2004fast}, semantic role
labeling~\citep{punyakanok2008importance,srikumar2011joint}, dependency
parsing~\citep{riedel2006incremental,martins2009concise}, coreference
resolution~\citep{denis2009global}, sentence
compression~\citep{clarke2008global,thadani2013sentence}, inferring
alignments~\citep{goldwasser2008transliteration,chang2010discriminative,li2016exploiting},
summarization~\citep{woodsend2012multiple},
supertagging~\citep{ravi2010minimized}, common sense
reasoning~\citep{roy2015solving,goldwasser2016understanding}, and many others.
It is important to point out that these examples include both cases where the
computational aspects of inference were handled by powerful off-the-shelf
solvers such as Express-MP or Gurobi, and those where approximate methods were
designed for inference.\footnote{See, for example,
  \url{https://ilpinference.github.io/eacl2017/} for details.}

The integer linear programming formalism is both expressive and easy to use for
representing and reasoning with knowledge for two reasons. First, every MAP
inference problem with discrete variables 
can be represented as a linear objective~\citep{roth2007global}, making ILP a
natural formalism for such problems. Second, all Boolean functions can be
compiled into a set of linear inequalities, to be used as constraints in the ILP
formulation.

This tutorial-style survey paper focuses on this second point, and is meant to
guide the reader through the process of framing a new inference problem as an
instance of an integer linear program. It is structured as a collection of
commonly used recipes, and at the end, we will see two worked examples to
illustrate the use of these recipes.

To simplify discourse, we will make two assumptions. First, we will assume that
we have all the scoring functions needed to write the objective
function. Second, we will primarily focus on the process of writing down the
inference problems, not solving them. It is important to separate the
declaration of a problem from its solution; this article concerns the former.
We could solve inference problems using off-the-shelf black box solvers, general
heuristics, or specially crafted algorithms tailored to the problem at hand.

A final note before we get started: While the motivating examples used in this
paper are drawn from natural language processing, the techniques for converting
Boolean expressions into linear inequalities that are discussed here are applicable more
broadly. As a result, the next few sections are written
without a specific domain in mind, but the worked examples that follow are
grounded in NLP tasks\@.


\section{Notation and Preliminaries}
\label{sec:notation}

To start off, let us first see the notation that will be used
through this survey.

\paragraph{Decision variables.} Our goal is to collectively make a
set of possibly interacting decisions. We will refer to individual
Boolean decisions using the symbol $\var{}$ with
subscripts. Usually, the decisions in the subscripts deal with
assigning labels to inputs. For example, the decision that the
$i^{th}$ label is \lbl{A} will be represented as
$\varlab{i}{\lbl{A}}$. For brevity, if the label $A$ is the constant
\true, we will write $\var{i}$ to denote $\varlab{i}{\true}$.

We can map from the space of Boolean decisions (i.e., predicates) to
integers using the Iverson bracket~\citep{iverson1962programming}.
The Iverson bracket for a predicate $\var{}$, denoted by $\ivar{}$, is
defined as
\begin{equation}
  \label{eq:iversion}
  \ivar{} = \begin{cases}
    1 & \mbox{if }\var{}\mbox{ is \true} \\
    0 & \mbox{if }\var{}\mbox{ is \false}.
   \end{cases}
\end{equation}
In other words, it maps \true{} to $1$ and \false{} to $0$.  As
\citet{knuth1992two} points out, the Iverson bracket is a notational
convenience that vastly simplifies mathematical exposition.
Here, we will assume the implicit existence of the Iverson bracket to
translate \false\ and \true\ to $0$ and $1$ respectively. This implicit
notational device will allow us to reason about Boolean variables like
$\var{}$ as if they were integers.

Each decision $\var{i}$ is associated with a score $\coeff{i}$. We
will assume the convention that we prefer decisions whose scores are
larger. Importantly, in this survey, we will not concern ourselves
with where the scores originate; the scoring function could have
been learned in the past, or the inference could be situated within
the context of a learning algorithm that estimates the scoring
function, or perhaps the scores were manually set using domain
knowledge. Furthermore, we do not make any assumptions about the
nature of the scores---while they could represent log probabilities
that the corresponding variable is \true, we do not assume that they
are probabilities in the formal sense; we merely require that
variable assignments that are associated with a higher total score are preferable.

Finally, we will use the boldface symbol $\vars$ to denote a vector of decision
variables and the boldface $\coeffs$ to denote the vector of coefficients that
score the decision variables in $\vars$.

\paragraph{Integer linear programs.} The goal of inference is to assign
values to the decision variables such that their total score is
maximized. We will formalize this task as an integer linear program
(ILP).  To define the integer linear program, we need to specify a
linear objective function and a collection of linear constraints that
characterize the set of valid decisions. In general, we can write the
inference problem as
\begin{eqnarray}
  \max_{\vars} & \Sum_i \coeff{i}\var{i} \label{eq:general-ilp:obj} \\
  \mbox{s.t. } & \vars \in \sY,          \label{eq:general-ilp:constraints} \\
               & \var{i} \in \{0, 1\}.  \label{eq:general-ilp:0-1}
\end{eqnarray}
Here, $\sY$ denotes a set of legal assignments to the inference
variables. The actual definition of this set in the form of linear
inequalities is dependent on the problem and the subsequent sections
are devoted to recipes for constructing this set.

Of course, even the definition of the inference variables is a
problem-specific design choice.  The inference variables in the
objective function are constrained to be zero or one. Thus, our
problem is an instance of a $0$-$1$ integer linear program. The linear
objective \eqref{eq:general-ilp:obj} ensures that only the
coefficients for variables that are assigned to \true (or
equivalently, to 1 via the Iverson bracket) will count towards the total
score. While not explicitly stated in the formulation above, we can 
also add additional auxiliary discrete or real valued inference
variables to allow us to state the problems in an easy way or to
facilitate solving them.

Integer and mixed-integer programming is well studied in the
combinatorial optimization literature. An overview of their
computational properties is beyond the scope of this survey and the
reader should refer to textbooks that cover this topic~\cite[for
example]{papadimitriou1982combinatorial,schrijver1998theory}.  For our purposes, we
should bear in mind that, in general, integer programming is an
NP-hard problem. Indeed, $0$-$1$ integer programming was one of
Karp's 21 NP-complete problems~\citep{karp1972complexity}. Thus, while the
techniques described in this tutorial provide the tools to encode
our problem as an integer program, we should be aware that we may
end up with a problem formulation that is intractable. For certain
NLP problems such as semantic role labeling~\cite[for
example]{punyakanok2008importance}, we can show that certain ways to model
the problem leads to inference formulations that are intractable in
the worst case. Yet, curiously, in practice, off-the-shelf solvers
seem to solve them quite fast! Indeed, the same problem could be
encoded in different ways, one of which can be solved efficiently
while another is not. One example of this situation is the task of
graph-based dependency parsing. The ILP encoding of
\citet{riedel2006incremental} required a specialized cutting-plane method,
while the flow-inspired encoding of \citet{martins2009concise} was more
efficiently solvable.


\section{Basic Operators: Logical Functions}
\label{sec:basic-operators}

In this section, we will introduce the basic building blocks needed to
convert Boolean expressions into a set of linear inequalities. For now, we will only use $0$-$1$ decision
variables as described in \S\ref{sec:notation} without any auxiliary real-valued
variables. Using {\em only} the techniques described in this section,
we should be able to write any Boolean expression as a set of
linear inequalities.

\subsection{Variables and their Negations}
\label{sec:variables-negations}
Recall that each variable $\var{}$ in the $0$-$1$ ILP corresponds to a
Boolean decision. A natural first constraint may seek to enforce a
certain set of decisions, or equivalently, enforce their logical
conjunction. This gives us our first recipe.

\begin{constraint}{Forcing the conjunction of decisions $\var{1}, \var{2}, \ldots, \var{n}$ to be
    \true. That is, $\var{1}\land\var{2}\land\cdots\land\var{n}$.}
    \Sum_{i=1}^n\var{i} = n.
\end{constraint}
Since the decision variables can only be $0$ or $1$, the sum in the
constraint counts the number of decisions enforced. With $n$
variables, this sum can be $n$ if, and only if, each one of them
takes the value $1$.

\paragraph{Handling negations.} Setting a variable $\var{}$ to \false\
is equivalent to setting $1 - \var{}$ to \true. This observation
gives us a general strategy to deal with negations: Suppose a variable $\var{}$ is negated in a Boolean
expression. While converting this expression into a linear
inequality (using one of the recipes in this survey), we will
replace occurrences of $\var{}$ in the inequality with $1 -
\var{}$. For example, the constraint $\neg\var{}$ would become
$1 - \var{} = 1$ (or $\var{} = 0$).
Applying this strategy to the above constraint gives us a second
constraint that forbids a collection of $n$ decisions from being
\true.
\begin{constraint}{Forbidding all the decisions
    $\var{1}, \var{2}, \ldots, \var{n}$ from being \true. That is,
    $\neg\var{1} \wedge\neg\var{2}\wedge\cdots\wedge\neg\var{n}$.}
  \Sum_{i=1}^n \var{i} = 0.
\end{constraint}
The need to force decision variables to be either \true or \false
arises when we wish to unconditionally enforce some external knowledge
about the prediction.

\begin{example}
  Suppose we know the ground truth assignments for a subset of our
  decision variables and we wish to ascertain the best assignment to
  the other variables according to our model. We could do so by
  forcing the known variables to their values. Such an approach
  could be useful for training models with partial supervision over
  structures.
\end{example}

\begin{example}[Testing inference formulations]
  Another use case for the above constraint recipes is that it
  offers a way to check if our inference formulation for a problem is correct. Suppose we have
  a labeled data set that maps inputs $\bx$ (e.g., sentences) to
  outputs $\by$ (e.g., labeled graphs) and we have framed the
  problem of predicting these outputs as an ILP.

  One way to test whether our problem formulation (as defined by our
  constraints) is meaningful is to add additional constraints that
  clamp the decision variables to their ground truth labels in a
  training set. If the resulting ILP is infeasible for any example,
  we know that the \emph{rest} of our constraints do not accurately
  reflect the training data. Of course, we may choose not to correct
  this inconsistency with the data, but that is a modeling choice.
\end{example}

\subsection{Disjunctions and their Variants}
\label{sec:disjunctions}
An important building block in our endeavor is the disjunction.
Suppose we have a collection of decision variables and we require that
at least one of them should hold.  Using the Iverson notation
naturally gives us the constraint formulation below.

\begin{constraint}{Disjunction of $\var{1}, \var{2}, \ldots, \var{n}$.
    That is, $\var{1}\vee\var{2}\vee\cdots\vee\var{n}$.}
  \Sum_{i=1}^n\var{i} \geq 1
\end{constraint}

Note that this constraint can incorporate negations using the
construction from \S\ref{sec:variables-negations}, as in the
following example.

\begin{example}
  If we want to impose the constraint
  $\neg\var{1}\vee\neg\var{2}\vee\neg\var{3}$, we need to use
  $1 - \var{1}$, $1-\var{2}$ and $1-\var{3}$ in the recipe above. This
  gives us
  \begin{align*}
                     & 1 - \var{1} + 1 - \var{2} + 1 - \var{3}  \geq  1, \\
    \mbox{that is, } & \var{1} + \var{2} + \var{3} \leq  2.
  \end{align*}
\end{example}

There are several variations on this theme. Sometimes, we may require
that the number of \true assignments should be at least, or at most,
or exactly equal to some number $k$. These {\em counting quantifiers}
or {\em cardinality quantifiers} generalize both conjunctions and
disjunctions of decisions. A conjunction of $n$ variables demands that
the number of \true assignments should be equal to $n$; their
disjunction demands that at least one of the variables involved should be \true.

\begin{constraint}{At least, at most or exactly $k$ \true assignments
    among $\var{1}, \var{2}, \ldots, \var{n}$.}
  \mbox{At least $k$:} & \Sum_{i=1}^n\var{i} \geq k \\
  \mbox{At most $k$:}  & \Sum_{i=1}^n\var{i} \leq k \\
  \mbox{Exactly $k$:}  & \Sum_{i=1}^n\var{i}  = k.
\end{constraint}

The use of counting quantifiers does not increase the expressive
power over logical expressions. They merely serve as a syntactic
shorthand for much larger Boolean expressions. For example, if we
wish to state that exactly two of the three variables $\var{1}$,
$\var{2}$ and $\var{3}$ are \true, we can encode it using the
following expression:
\begin{equation*}
      \p{\var{1}    \land\var{2}    \land\neg\var{3}}
  \lor\p{\var{1}    \land\neg\var{2}\land\var{3}}
  \lor\p{\neg\var{1}\land\var{2}    \land\var{3}}
\end{equation*}

An important (that is, frequently applicable) special case of counting
quantifiers is {\em uniqueness quantification}, where we require
exactly one of a collection of decisions to hold. While the
corresponding linear constraint is clearly easy to write using what we
have seen above, uniqueness constraints are important enough to merit
stating explicitly.

\begin{constraint}{Unique assignment among
    $\var{1},\var{2},\ldots,\var{n}$. That is, $\exists !~\var{i}$.}
  \Sum_{i=1}^n \var{i} = 1.
\end{constraint}

As an aside, this constraint is identical to the logical XOR if we have exactly
two variables (i.e., their parity is one when the constraint holds), but not
when the number of variables is more. For example, with three variables, if all
of them are assigned to \true, their parity is one, but the above constraint is
not satisfied.

\begin{example}[Multiclass classification]
  The linear constraint templates described in this section find wide
  applicability. The simplest (albeit unwieldy)
  application uses the unique label constraint to formally define
  multiclass classification. Suppose we have inputs that are to be
  assigned one of $n$ labels
  $\{\lbl{l_1}, \lbl{l_2}, \ldots, \lbl{l_n}\}$. We can write this
  prediction problem as an integer linear program as follows:
  \begin{eqnarray*}
    \max_{\vars}      & \Sum_{i=1}^n \coefflab{\labelString}{\lbl{l_i}}\cdot\varlab{\labelString}{\lbl{l_i}} \\
    \mbox{such that } & \Sum_{i=1}^n \varlab{\labelString}{\lbl{l_i}} = 1,                              \\
                      & \varlab{\labelString}{\lbl{l_i}} \in \{0,1\}.                                   \\
  \end{eqnarray*}
  We have $n$ decision variables, each corresponding to one of the
  possible label assignments. The decision of choosing the label
  $\lbl{l_i}$ is scored in the objective function by a score
  $\coefflab{\labelString}{\lbl{l_i}}$. The goal of inference is to
  find the score maximizing assignment of these decision
  variables. The constraint mandates that exactly one of the inference
  outcomes is allowed, thus ensuring that the label that maximizes the
  score is chosen.
\end{example}
The above example merely illustrates the use of the unique label
constraint. While inference for multiclass classification can be written in this
form, it is important to note that it is unwise to use a black box ILP solver to
solve it; simply enumerating the labels and picking the highest scoring one
suffices. This example highlights the difference between {\em framing} a problem
as an integer linear program and {\em solving} it as one. While multiclass
classification can clearly be framed as an ILP, solving it as one is not a good
idea.

However, the {\sc multiclass as an ILP} construction is a key building block for
defining larger structured outputs. A commonly seen inference situation requires
us to a unique label to each of a collection of categorical random variables,
subject to other constraints that define the interactions between them. In such
a situation, each categorical random variable will invoke the {\sc multiclass as
  an ILP} construction.


\subsection{A recipe for Boolean expressions}
\label{sec:recipe-bool-expressions}

In \S\ref{sec:variables-negations} and \S\ref{sec:disjunctions}, we
saw recipes for writing Boolean variables, their negations,
conjunctions and disjunctions as linear inequalities. With the full complement
of operators, we can convert any  constraint
represented as a Boolean expression into a collection of linear
inequalities using the following procedure:
\begin{enumerate}
\item Convert the Boolean expression into its conjunctive normal
  form (CNF) using De Morgan's laws and the distributive property,
  or by introducing new variables and using the Tseitin
  transformation~\citep{tseitin1983complexity}.
\item Recall that a CNF is a conjunction of disjunctive
  clauses. Express each clause in the CNF (a disjunction) as a
  linear inequality.
\end{enumerate}

Let us work through this procedure with two examples. In both
examples, we will not worry about the objective function of the ILP
and only deal with converting Boolean expressions into linear
constraints.
\begin{example}
  Suppose we have three Boolean variables $\var{1}$, $\var{2}$ and
  $\var{3}$ and our goal is to convert the following Boolean
  expression into linear inequalities:
  \begin{equation*}
    \p{\var{1}\wedge\neg\var{2}} \vee\p{\var{1}\wedge\neg\var{3}} 
  \end{equation*}
  The first step, according to the recipe above, is to convert this
  into its equivalent conjunctive normal form:
  \begin{equation*}
    \var{1}\wedge%
    \p{\neg\var{2}\vee\neg\var{3}}.
  \end{equation*}
  Now, we have two clauses, each of which will become a linear
  constraint. Using the templates we have seen so far and
  simplifying, we get the following linear constraints:
  \begin{eqnarray*}
    \var{1} = 1,\\
    \var{2} + \var{3} \leq 1.
  \end{eqnarray*}
\end{example}

\begin{example}\label{eg:xor}
  Suppose we have three decision variables $\var{1},\var{2}$ and
  $\var{3}$ and we wish to enforce the constraint that either all of
  them should be \true or all of them should be \false.  The
  constraint can be naturally stated as:
  \begin{equation*}
    \p{\var{1} \wedge \var{2} \wedge \var{3}} \vee%
    \p{\neg\var{1} \wedge \neg\var{2} \wedge \neg\var{3}}.
  \end{equation*}
  To express the constraint as a set of linear inequalities, let us
  first write down its conjunctive normal form:
  \begin{equation*}
    \p{\var{1}\vee\neg\var{2}} \wedge%
    \p{\var{1}\vee\neg\var{3}} \wedge%
    \p{\var{2}\vee\neg\var{1}} \wedge%
    \p{\var{2}\vee\neg\var{3}} \wedge%
    \p{\var{3}\vee\neg\var{1}} \wedge%
    \p{\var{3}\vee\neg\var{2}}.
  \end{equation*}
  Now, we can convert each disjunctive clause in the CNF form to a
  different linear constraint following the templates we have seen
  before. After simplification, we get the following linear system
  that defines the feasible set of assignments:
  \begin{eqnarray*}
    \var{1} - \var{2} \geq 0,\\
    \var{1} - \var{3} \geq 0, \\
    \var{2} - \var{1} \geq 0, \\
    \var{2} - \var{3} \geq 0, \\
    \var{3} - \var{1} \geq 0, \\
    \var{3} - \var{2} \geq 0.
  \end{eqnarray*}
\end{example}

The procedure provides a systematic approach for converting Boolean
constraints (which are easier to state) to linear inequalities
(allowing us to use industrial strength solvers for probabilistic
inference). Indeed, the recipe is the approach suggested by
\citet{rizzolo2007modeling} and \citet{rizzolo2012learning} for learning based
programming. However, if applied \naively, this methodical approach
can present us with difficulties with respect to the number of linear
constraints generated.

Consider the final set of inequalities obtained in Example~\ref{eg:xor} above.
While we could leave the linear system as it is, the system of
equations implies that $\var{1} = \var{2} = \var{3}$, as does the
logical expression that we started with. This example illustrates an
important deficiency of the systematic approach for converting logical
formulae into linear inequalities. While the method is sound and
complete, it can lead to a much larger set of constraints than
necessary. 
We will see in \S\ref{sec:implications} that such ``improperly'' encoded
constraints can slow down inference.

One way to address such a blowup in the number of constraints is to
identify special cases that represent frequently seen inference
situations and lead to large number of constraints, and try to find
more efficient conversion techniques for them. The following sections
enumerate such special cases, starting with implications
(\S\ref{sec:implications}) and moving on to combinatorial structures
(\S\ref{sec:complex-building-blocks}).



\section{Simple and Complex Logical Implications}
\label{sec:implications}

The first special case of constraints we will encounter are conditional
forms. At first, we will simply convert the implications into disjunctions and
use the disjunction templates from \S\ref{sec:basic-operators}. Then, in
\S\ref{sec:complex-implications}, we will exploit the fact that our inference
variables can only be $0$ or $1$ to reduce the number of constraints. 

\subsection{Simple Conditional Forms}
\label{sec:simple-implications}

First, let us consider the simplest implication constraint:
$\var{1} \rightarrow \var{2}$. Clearly, this is equivalent to the
disjunction $\neg\var{1}\vee\var{2}$ and we can convert it to the
constraint $-\var{1} + \var{2} \geq 0$. 
We can generalize this to a conditional form with a conjunctive
antecedent and a disjunctive consequent:
\begin{equation*}
  \Wedge_{i=1}^m \var{l_i} \rightarrow \Vee_{i=1}^{n} \var{r_i}.
\end{equation*}
The implication is equivalent to the disjunction:
\begin{equation*}
  \p{\Vee_{i=1}^m \neg\var{l_i}} \bigvee \p{\Vee_{i=1}^n \var{r_i}}.
\end{equation*}
Now, we can use the disjunction and negation rules that we have seen
before. We get
\begin{eqnarray*}
  \Sum_{i=1}^m \p{1 - \var{l_i}} + \Sum_{i=1}^{n} \var{r_i} \geq 1.
\end{eqnarray*}
Simplifying the expression and moving constants to the right hand side
gives us our next recipe:
\begin{constraint}{Implications of the form
    $\Wedge_{i=1}^m \var{l_i} \rightarrow \Vee_{i=1}^{n} \var{r_i}$}
  -\Sum_{i=1}^m \var{l_i} + \Sum_{i=1}^{n} \var{r_i} \geq 1 - m.
\end{constraint}
One special case merits explicit mention---the Horn clause, which is
well studied in logic programming~\citep{chandra1985horn}.
\begin{constraint}{Horn clauses of the form
    $\var{l_1}\wedge\var{l_2}\wedge\cdots\wedge\var{l_m}\rightarrow\var{r}$}
  -\Sum_{i=1}^m \var{l_i} + \var{r} \geq 1 - m.
\end{constraint}
%


\subsection{Complex conditional forms}
\label{sec:complex-implications}
Suppose we have three decisions $\var{1}$, $\var{2}$ and $\var{3}$ and
we require that the decision $\var{3}$ holds if, and only if, both
$\var{1}$ and $\var{2}$ hold. We can write this requirement as
\begin{equation}
  \var{1} \wedge \var{2} \leftrightarrow \var{3}.\label{eq:complex-conditional:example} 
\end{equation}
The constraint can be written as two implications:
\begin{eqnarray}
  \var{1} \wedge \var{2} \rightarrow \var{3} \label{eq:complex-conditional:right}\\
  \var{3} \rightarrow \var{1} \wedge \var{2}. \label{eq:complex-conditional:left}
\end{eqnarray}
The first implication matches the template we saw in
\S\ref{sec:simple-implications} and we can write it as
$-\var{1} -\var{2} + \var{3} \geq -1$. The second one can be broken down into
two conditions $\var{3} \rightarrow \var{1}$ and $\var{3} \rightarrow
\var{2}$. These correspond to the inequalities $\var{1} - \var{3} \geq 0$ and
$\var{2} -\var{3} \geq 0$ respectively.  In other words, the single
biconditional form, following the methodical approach, gets translated into
three linear inequalities.  In general, if there are $n$ elements in the
conjunction on the left hand side of the implication, we will have $n + 1$
linear inequalities. Can we do better?\footnote{We should point out that we are
  working under the assumption that fewer, more dense inequalities are better
  for solvers. Indeed, the experiments in \S\ref{sec:empirical} corroborate this
  assumption. However, while seems to empirically hold for solvers today, the
  inner workings of a solver may render such optimization unnecessary.}

In this section, we will see several commonly seen design patterns
concerning conditional expressions. It summarizes and generalizes
techniques for converting conditional forms into linear inequalities
from various sources~\cite[inter
alia]{gueret2002applications,punyakanok2008importance,noessner2013rockit}.

\paragraph{Equivalence of decisions.} 
Suppose we wish to enforce that two decision variables should take the
same value. If this condition were written as a logical expression, we
would have $\var{1} \leftrightarrow \var{2}$. We saw in the example in
\S~\ref{sec:recipe-bool-expressions} that \naively\ converting the
implication into a CNF and proceeding with the conversion leads to two
constraints per equivalence. Instead, we can use the facts that the
decisions map to numbers, and that we have the ability to use linear equations,
and not just inequalities, to get the following natural constraint:
\begin{constraint}{Equivalence of two variables:
    $\var{1}\leftrightarrow \var{2}$.}
  \var{1} - \var{2} = 0.
\end{constraint}

\paragraph{Disjunctive Implication.} Suppose we have two collections
of inference variables $\var{l_1}$, $\var{l_2}$,$\cdots$,
$\var{l_n}$ and $\var{r_1}, \var{r_2}, \cdots, \var{r_m}$. We wish
to enforce the constraint that if {\em any} of the $\var{l_i}$
decisions are \true, then at least one of the $\var{r_i}$'s should
be \true. It is easy to verify that if written \naively, this will
lead to $n$ linear inequalities. However, only one suffices.

\begin{constraint}{Disjunctive Implication:
    $\Vee_{i=1}^n\var{l_i}\rightarrow\Vee_{i=1}^m\var{r_i}$}
  -\Sum_{i=1}^n\var{l_i} + n\Sum_{i=1}^m \var{r_i} \geq 0.
\end{constraint}

To show that this is correct, let us consider two cases.

\begin{enumerate}
\item First, if the left hand side of the implication is \false\ (i.e., none of
  the $\var{l_i}$'s are \true), then the implication holds. In this case, we see
  that the inequality is satisfied no as negative terms remain on its left hand
  side.

\item Second, if the left hand side of the implication is \true, then at least
  one, and as many as $n$ of the $\var{l_i}$'s are \true. Consequently, the sum
  of the negative terms in the inequality can be as low as $-n$. For the
  implication to hold, at least one of the $\var{r_i}$'s should be \true. But if
  so, we have $n\sum\var{r_i} \geq n$. In other words, the left hand side of the
  inequality becomes non-negative.
\end{enumerate}

We see that the inequality is satisfied whenever the implication holds.

\paragraph{Conjunctive Implication.} This setting is similar to the
previous one. We have two collections of inference variables
$\var{l_1}, \var{l_2},\cdots,\var{l_n}$ and
$\var{r_1}, \var{r_2}, \cdots, \var{r_m}$. We wish to enforce the
constraint that if {\em all} the $\var{l_i}$'s are \true, then {\em
  all} the $\var{r_i}$'s should be \true. As with the case of disjunctive implications, if written
\naively, this will lead to $m$ linear inequalities. 
Once again, we can compactly encode the requirement with one
inequality.
\begin{constraint}{Conjunctive implication:
    $\Wedge_{i=1}^n\var{l_i}\rightarrow\Wedge_{i=1}^m\var{r_i}$}     
   -m\Sum_{i=1}^n \var{l_i} + \Sum_{i=1}^m \var{r_i} \geq m(1 -n).
\end{constraint}
Intuitively, if even one of the $\var{l_i}$'s is \false, the
inequality holds irrespective of the number of $\var{r_i}$'s that are
true. However, if all the $\var{l_i}$'s are \true, then every
$\var{r_i}$ needs to be \true for the inequality to hold.
To show the correctness the above recipe, consider the contrapositive
of the conjunctive implication:
$\Vee_{i=1}^m\neg\var{r_i}\rightarrow\Vee_{i=1}^n\neg\var{l_i}$. We
have a disjunctive implication where all variables are negated. We can
use the recipe for disjunctive implications from above, but replace
all variables $\var{l_i}$ and $\var{r_i}$ with $1 - \var{l_i}$ and
$1 - \var{r_i}$ to account for the fact that they are
negated. Cleaning up the resulting inequality gives us the recipe for
conjunctive implications. 

\begin{example}
  Using the conjunctive implication, we can now revisit the
  constraint~\eqref{eq:complex-conditional:example} we saw at the beginning of
  this section, and see that it can be written using only two inequalities
  instead of three. As earlier, we will write this biconditional form as two
  conditional forms \eqref{eq:complex-conditional:left} and
  \eqref{eq:complex-conditional:right}. The first one, being a simple
  conditional form, corresponds to one constraint. The second constraint
  $\var{3} \rightarrow \var{1} \wedge \var{2}$ is a conjunctive implication and
  can be written as the single inequality
  $-2\var{3} + \var{1} + \var{2} \geq 0$.
\end{example}

Clearly other conditional forms that are not discussed here are
possible. However, not all of them are amenable to being reduced to a
single inequality. The usual strategy to handle such complex
conditional forms is to symbolically transform a constraint
into the forms described here and convert the resulting constraints
into a system of linear inequalities.

Complex implications are useful to write down many-to-many
correspondences between inference assignments.  The need to write down
many-to-many correspondences arises naturally when we are predicting
labels for nodes and edges of a graph and we wish to restrict values
of edge labels based on the labels assigned to nodes to which the edge
is incident.

\begin{example}\label{eg:complex-implication-example}
  To illustrate an application of complex implications, consider a
  problem where we have a collection of slots, denoted by the set
  $\sS = \{S_1, S_2, S_3, \cdots\}$. Suppose our goal is to assign a
  unique label from
  $\sL = \{\lbl{l_1}, \lbl{l_2}, \lbl{l_3}, \lbl{l_4}\}$ to each
  slot.

  The problem definition naturally gives us inference variables of the
  form $\varlab{S_i}{\lbl{l_j}}$ that states that the slot $S_i$ is
  assigned a label $\lbl{l_j}$. The uniqueness constraint can be
  written as a Boolean expression demanding that, for every slot,
  there is a unique label. 
  \begin{equation*}
    \forall s \in \sS, \quad \exists !~\lbl{l} \in \sL, \varlab{s}{\lbl{l}}.
  \end{equation*}
  We can write this constraint as a collection of linear inequalities,
  using the {\sc multiclass as an ILP} construction:
  \begin{equation*}
    \forall s \in \sS, \quad \Sum_{l \in \sL} \varlab{s}{\lbl{l}} = 1.
  \end{equation*}
  
  In addition, suppose our knowledge of the task informs us that the
  slots $S_1$ and $S_2$ constrain each other:
  \begin{quote}
    The slot $S_1$ can assume one of the labels $\lbl{l_1}$ or $\lbl{l_2}$ if,
    and only if, the slot $S_2$ is assigned either the label $\lbl{l_3}$ or
    $\lbl{l_4}$.

    Likewise, $S_1$ can assume one of $\lbl{l_3}$ or $\lbl{l_4}$ if, and only
    if, the slot $S_4$ is assigned either the label $\lbl{l_1}$ or $\lbl{l_2}$.
  \end{quote}
  This domain knowledge can be formally written as
  \begin{align*}
    \varlab{S_1}{\lbl{l_1}} \vee \varlab{S_1}{\lbl{l_2}} & \leftrightarrow \varlab{S_2}{\lbl{l_3}} \vee \varlab{S_2}{\lbl{l_4}},  \\
    \varlab{S_4}{\lbl{l_1}} \vee \varlab{S_4}{\lbl{l_2}} & \leftrightarrow \varlab{S_1}{\lbl{l_3}} \vee \varlab{S_1}{\lbl{l_4}}.
  \end{align*}
  Each constraint here is a biconditional form, which can be written
  as two disjunctive implications and subsequently converted into
  linear inequalities using the recipe we have seen earlier in this
  section:
  \begin{align*}
    - \varlab{S_1}{\lbl{l_1}} -  \varlab{S_1}{\lbl{l_2}} + 2 \varlab{S_2}{\lbl{l_3}}  +  2\varlab{S_2}{\lbl{l_4}} & \geq 0, \\
    - \varlab{S_2}{\lbl{l_3}} -  \varlab{S_2}{\lbl{l_4}} + 2 \varlab{S_1}{\lbl{l_1}} +  2\varlab{S_1}{\lbl{l_2}}  & \geq 0, \\
    - \varlab{S_4}{\lbl{l_1}} -  \varlab{S_4}{\lbl{l_2}} + 2 \varlab{S_1}{\lbl{l_3}}  +  2\varlab{S_1}{\lbl{l_4}} & \geq 0,\\
    - \varlab{S_1}{\lbl{l_3}} -  \varlab{S_1}{\lbl{l_4}} + 2 \varlab{S_4}{\lbl{l_1}} +  2\varlab{S_4}{\lbl{l_2}}  & \geq 0.
  \end{align*}
  It should be easy to verify that if we had used Boolean operations to convert
  each of the biconditional forms into a conjunctive normal form and then
  applied the recipes from \S \ref{sec:basic-operators}, we would end up with
  eight inequalities instead of the four listed above.
\end{example}


\subsection{The Case for Special Cases: Empirical Evidence}
\label{sec:empirical}

The above discussion assumes that fewer inequalities are better handled by
solvers.  To see that this is indeed the case, let us look at the results of
experiments where we compare the \naive\ conversion of conjunctive and disjunctive
implications (i.e., via their conjunctive normal form, as in
\S\ref{sec:recipe-bool-expressions}), and their more compact counterparts
defined in this section.

We considered synthetic problems with $100$ categorical variables, each of which
can take $50$ values. As in Example~\ref{eg:complex-implication-example}, this
gives us $5000$ Boolean variables, with the unique label constraint within each
block. We constructed random implications of the form seen above using these
categorical variables, and their Boolean counterparts.  To do so, we sampled two
equally sized random sets of categorical variables to define the left- and
right- hand sides of the implication respectively, and assigned a random label
to each. Note that each label assignment gives us a Boolean inference
variable. We randomly negated half of these sampled inference variables and
constructed a conjunctive or disjunctive implication as per the experimental
condition.

Given above setup, the question we seek to resolve is: \emph{Is it more
  efficient to create a smaller number of compact inequalities than employing
  the \naive\ conversion approach via conjunctive normal forms?} We considered
two independent factors in our experiments: the number of implications, and the
fraction of categorical variables participating in one constraint, i.e., the
\emph{constraint density}. For different values of these factors, we constructed
$100$ integer linear programs using the both the \naive\ and complex conversion
strategies, and measured the average wall-clock time for finding a
solution.\footnote{All experiments were conducted on a 2.6 GHz Intel Core i5
  laptop using the Gurobi solver (\url{http://www.gurobi.com}), version 8.1. To
  control for any confounding effects caused by multi-core execution of the
  solver, we restricted the solver to use one of the machine's cores for all
  experiments.}

\begin{figure}
  \centering
  \begin{subfigure}[b]{0.45\textwidth}
    \includegraphics[width=\textwidth]{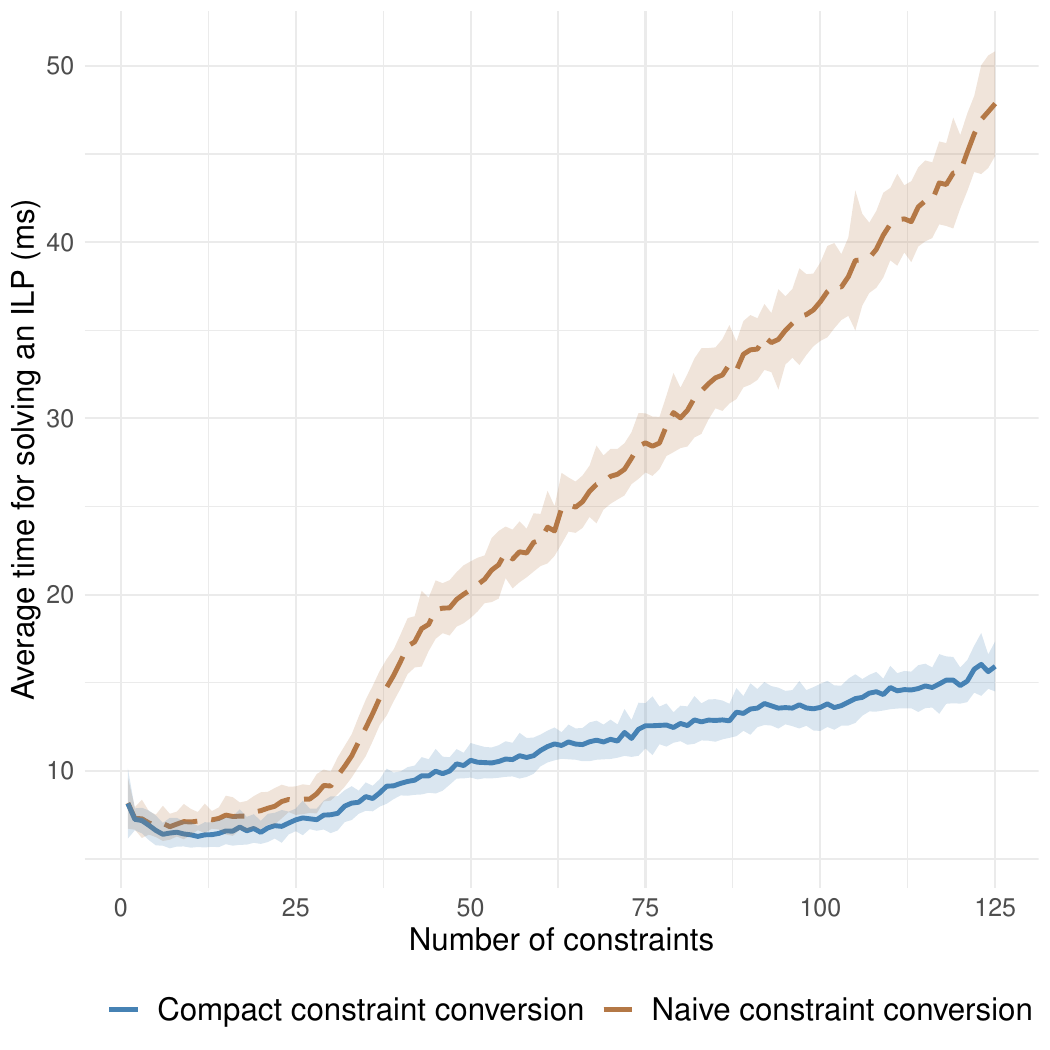}
    \caption{Constraint density = $20\%$}
    \label{fig:conjunctive-d-0.2}
  \end{subfigure}
  \hfill
  \begin{subfigure}[b]{0.48\textwidth}
    \includegraphics[width=\textwidth]{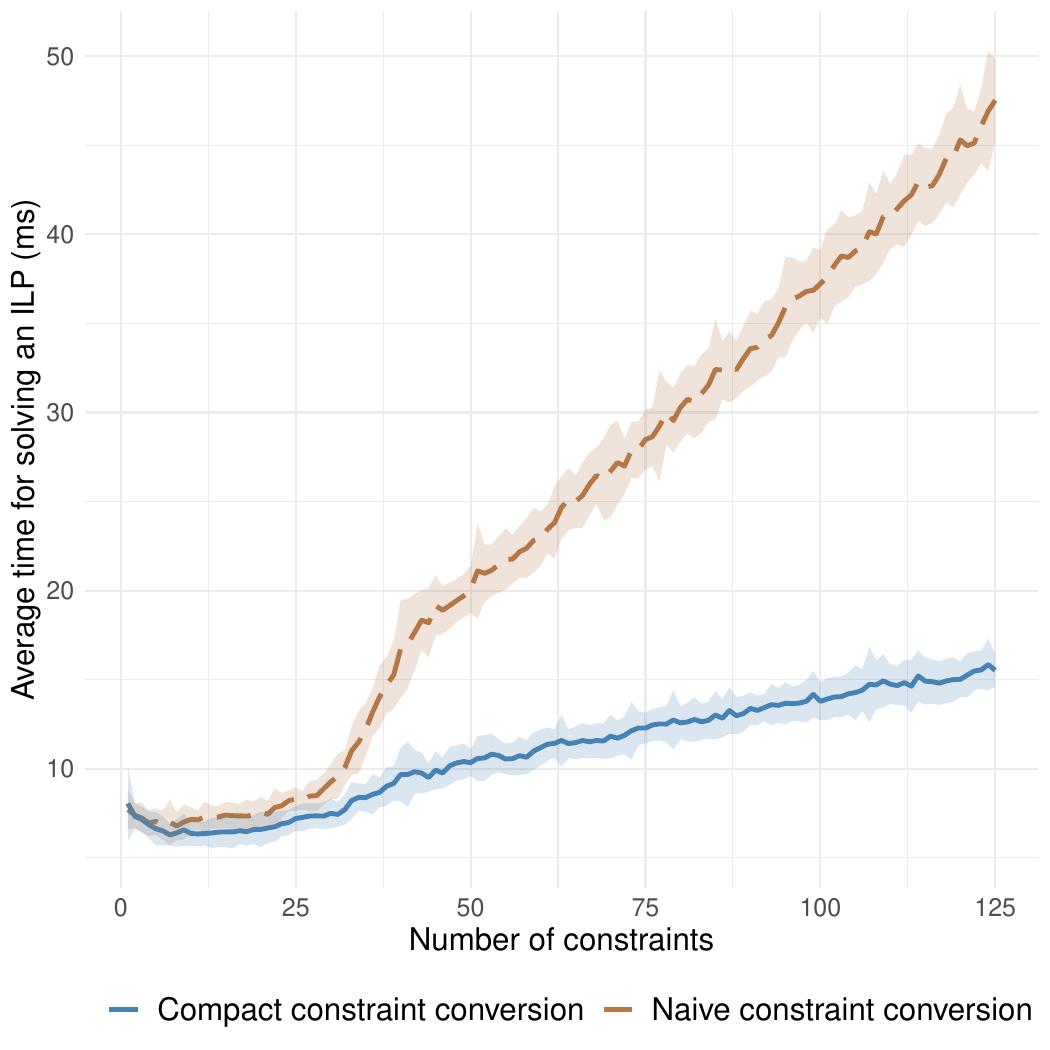}
    \caption{Constraint density = $40\%$}
    \label{fig:conjunctive-d-0.4}
  \end{subfigure}  
  \caption{Comparing encodings for conjunctive implications. The dashed brown
    lines show the average solver time (in milliseconds) across $100$ different
    runs for the \naive\ conversion to linear inequalities (\S\ref{sec:basic-operators}), while the solid blue lines
    correspond to the compact conversion (\S\ref{sec:implications}).
    The shaded regions show one standard deviation. The two subfigures show
    different constraint densities, which control how many categorical variables
    are involved in the implications. Across both conditions, the
    compact encoding is more efficient.}
  \label{fig:conjunctive-implication-comparison}
\end{figure}

\begin{figure}
  \centering
  \begin{subfigure}[b]{0.45\textwidth}
    \includegraphics[width=\textwidth]{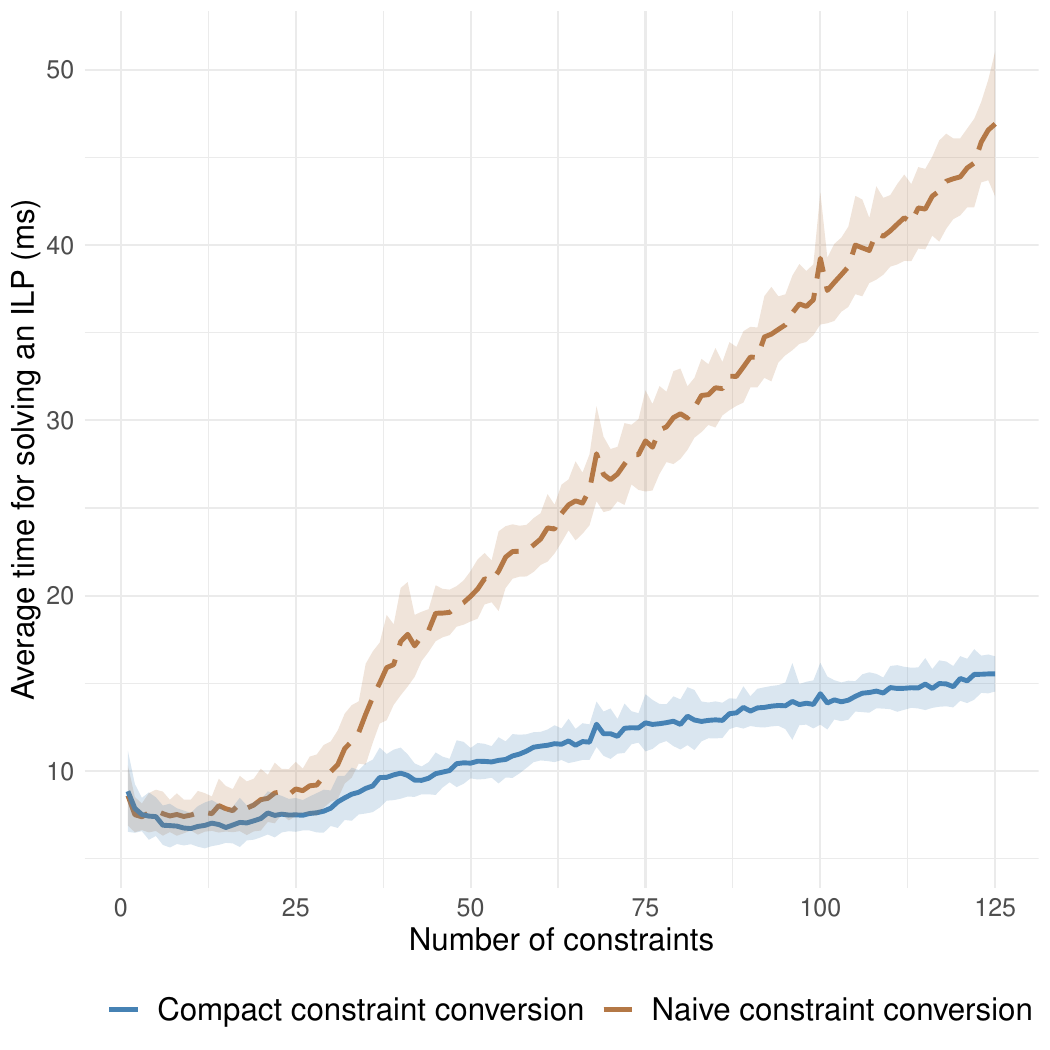}
    \caption{Constraint density = $20\%$}
    \label{fig:disjunctive-d-0.2}
  \end{subfigure}
  \hfill
  \begin{subfigure}[b]{0.45\textwidth}
    \includegraphics[width=\textwidth]{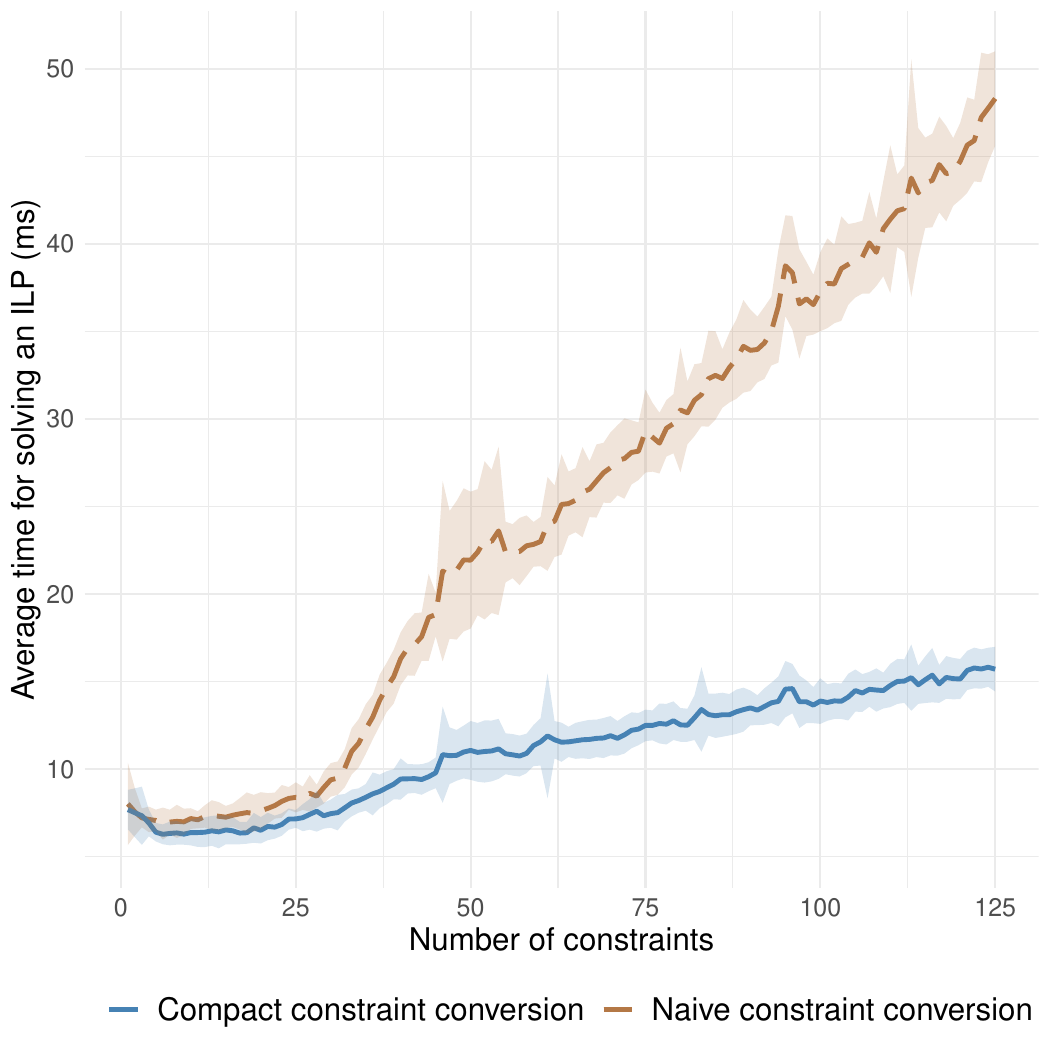}
    \caption{Constraint density = $40\%$}
    \label{fig:disjunctive-d-0.4}
  \end{subfigure}  
  \caption{Comparing encodings for disjunctive implications. See
    Figure~\ref{fig:conjunctive-implication-comparison} for details about the
    figure elements. As with conjunctions, compact encoding is
    more efficient.}
  \label{fig:disjunctive-implication-comparison}
\end{figure}

Figures~\ref{fig:conjunctive-implication-comparison}
and~\ref{fig:disjunctive-implication-comparison} show the results of these
experiments.  We see that for both kinds of implications, not only does the more
compact encoding lead to a solution faster, the time improvements increase as the
number of Boolean constraints increases.  Across all settings, we found that
when the number of Boolean constraints is over seven, the improvements in clock
time are statistically significant with $p < 0.001$ using the paired t-test.
These results show the impact of using fewer inequalities for encoding
constraints. For example, for conjunctive implications, with $100$ constraints,
we get over $2\times$ speedup in inference time. The results also suggest a
potential strategy for making a solver faster: if a solver could automatically
detect the inherent structure in the \naively\ generated constraints, it may be
able to rewrite constraints into the more efficient forms.

\section{Complex Building Blocks}
\label{sec:complex-building-blocks}

So far we have seen basic building blocks that can help us
declaratively construct output spaces for ILP inference. While any
Boolean expression can be expressed as linear inequalities using
only the tools introduced in \S\ref{sec:basic-operators}, we saw in
\S\ref{sec:implications} that certain Boolean predicates
(conditional forms) can be more compactly encoded as linear
inequalities than the \naive\ expansion would suggest.
In this section, we will look at more complex building blocks that
abstract away larger predicates efficiently. We will use the fact
that graph problems can be framed as linear programs to make these
abstractions. We demonstrate two inference situations that
frequently show up in NLP:\@ spanning tree constraints and graph
connectivity. We should note that other examples exist in the literature, for
example,~\citet{germann2004fast} studied the use of ILPs to define the decoding
problem  for machine translation as a traveling salesman problem. We refer the
reader to~\citet{trick2005formulations} for a discussion on using higher-order
constructs for constrained inference.

\paragraph{Notation.} Since we will be dealing with constraints on
graph structures, let us introduce the notation we will use for the
rest of this section. We will denote vertices of a graph by integers
$1, 2, \ldots, n$ and edges by pairs $(i,j)$. Thus, for any vertex $i$,
its outgoing edges are pairs of the form $(i, j)$ and incoming edges
are pairs of the form $(j,i)$.

\subsection{Spanning Trees}
\label{sec:spanning-trees}

Our first example concerns spanning trees. Suppose each edge in the
graph is associated with a score. Our goal is to identify the highest
scoring collection of edges that form a spanning tree. Of course,
efficient algorithms such as those of Bor\r{u}vka, Prim or Kruskal solve the
problem of finding maximum spanning trees for undirected graphs. If we
are dealing with directed graphs, then the equivalent problem of
finding the maximum spanning arborescence can be solved by the
Chu-Liu-Edmonds' algorithm.  However, we might want to enforce
additional task- or domain-specific constraints on the tree, rendering these
efficient maximum spanning tree (or arborescence) methods unsuitable.

To simplify discourse, we will assume that we have a fully connected,
undirected graph at hand. Our goal is to identify a subset of edges
that form a tree over the vertices. The construction outlined in this
section should be appropriately modified to suit variations.

Let us introduce a set of inference variables of the form $\var{ij}$
corresponding to an edge $(i,j)$ connecting vertices $i$ and $j$.
Since we are considering an undirected graph, and will not allow self-edges in
the spanning tree, we can assume that $i<j$ for all our inference variables.
If the variable $\var{ij}$ is set to \true, then the corresponding edge $(i,j)$ is selected
in the final sub-graph. One method for enforcing a tree structure is
to enumerate every possible cycle and add a constraint prohibiting
it. However, doing so can lead to an exponential number of
constraints, necessitating specialized solution strategies such as the
cutting plane method~\citep{riedel2006incremental}.

Alternatively, we can exploit the connection between network flow problems and
optimal trees to construct a more concise set of linear
inequalities~\citep{magnanti1995optimal,martins2009concise}. In particular, we
will use the well-studied relationship between the spanning tree problem and the
single commodity flow problem. In the latter, we are given a directed graph, and
we seek to maximize the total amount of a commodity (also called the flow)
transported from a source node to one or more target nodes in the graph. Each
edge in the graph has capacity constraints that limit how much flow it can
carry.

Without loss of generality, suppose we choose vertex $1$ to be root of
the tree. Then, we can write the requirement that the chosen vertices
should form a tree using the single commodity flow model as follows:
\begin{enumerate}
\item Vertex 1 sends a flow of $n-1$ units to the rest of the graph.
\item Each other vertex consumes one unit of flow. The amount of flow consumed by the
  node is simply the difference between its incoming and outgoing flows.
\item Only edges that are chosen to be in the tree can carry flow.
\end{enumerate}

To realize these three conditions, we will need to introduce auxiliary
non-negative integer (or real) valued variables $\flow{ij}$ and $\flow{ji}$ that
denote the flow associated with edge $(i,j)$ in either direction. Note that the
flow variables are directed even though the underlying graph is
undirected. These auxiliary variables do not feature in the ILP objective, or
equivalently they are associated with zero costs in the objective.

Using these auxiliary variables, we get the following recipe:

\begin{constraint}{Select a spanning tree among vertices $1,2,\cdots,n$ of a
    undirected graph using edge variables $\var{ij}$, where $i < j$. Introduce
    new integer variables $\flow{ij}$ and $\flow{ji}$ for every such pair
    $i,j$.}
  & \Sum_{j} \flow{1j} - \Sum_{j} \flow{j1}  =  n - 1, \\
  \mbox{for every vertex } i \in \{2,3,\cdots,n\}, & \Sum_{j} \flow{ji} - \Sum_{j}\flow{ij}  =  1,      \\
  \mbox{for every edge } (i,j),                    & \flow{ij} \leq (n-1)\var{ij},                      \\
  \mbox{for every edge } (i,j),                    & \flow{ji} \leq (n-1)\var{ij},                      \\
  \mbox{for every edge } (i,j),                    & \flow{ij} \geq 0,                                  \\
  \mbox{for every edge } (i,j),                    & \flow{ji} \geq 0,                                  \\
  & \Sum_{i,j}\var{ij} = n-1,                           
\end{constraint}

The first constraint here enforces that the chosen root sends a flow
of $n-1$ units to the rest of the vertices. The second one says that
every other vertex can consume exactly one unit of flow by mandating that the
difference between the total incoming flow and the total outgoing flow
for any vertex is 1. The third and fourth inequalities connect the
inference variables $\var{ij}$ to the flow variables by ensuring that
only edges that are selected (i.e. where $\var{ij}$ is \true) can
carry the flow. The next two constraints ensures that all
the flows are non-negative.  Finally, to ensure that the final
sub-graph is a tree, the last constraint ensures that exactly $n-1$
edges are chosen. We will refer these constraints collectively as the
{\sc Spanning Tree} constraints over the variables $\var{ij}$.

There are other ways to efficiently formulate spanning tree
constraints using linear inequalities. We refer the reader
to~\citet{magnanti1995optimal} for an extensive discussion involving tree
optimization problems and their connections to integer linear
programming.

To illustrate the {\sc Spanning Tree} construction, and how it can be used in
conjunction with other constraints, let us look at an example.

\begin{example}\label{eg:spanning-tree}
  Consider the graph in Figure~\ref{fig:flow-constraints-example}(a). Suppose
  our goal is to find a tree that spans all the nodes in the graph, and has the
  highest cumulative weight. To this end, we can instantiate the recipe detailed
  above.

  Each edge in the graph corresponds to one inference variable that determines
  whether the corresponding node is in the tree or not. The variables are
  weighted in the objective as per the edge weight. (We do not need to add
  variables for any edge not shown in the figure; they are weighted $-\infty$,
  and will never get selected.)  Collectively, all the edge variables, scaled by
  their corresponding weights, gives us the ILP objective to maximize, namely:
  \begin{align*}
                         10 \var{12} + 50 \var{13} + 5\var{15} + 11\var{23} + 15\var{15} -9 \var{34} - 7\var{35} - 50 \var{45}
  \end{align*}
  Next, we can instantiate the spanning tree constraints using flow variables
  $\{\flow{12}, \flow{21}, \cdots\}$. To avoid repetition, we will not rewrite
  the constraints here. Solving the (mixed) integer linear program with the flow
  constraints gives us an assignment to the $\var{ij}$ variables that
  corresponds to the tree in Figure~\ref{fig:flow-constraints-example}(b).  Of
  course, if our goal was merely to find the maximum spanning tree in the graph,
  we need not (and perhaps, should not) seek to do so via an ILP, and instead
  use one of the named greedy algorithms mentioned earlier that is specialized for this purpose.

  Now, suppose we wanted to find the second highest scoring tree. Such a
  situation may arise, for example, to find the top-$k$ solutions of an
  inference problem. To do so, we can add a single extra constraint in addition
  to the flow constraints that prohibit the tree from
  Figure~\ref{fig:flow-constraints-example} (b). In other words, the solution we
  seek should satisfy the following constraint:
  \begin{align*}
    \neg\left( \var{13} \land \var{23} \land \var{25} \land \var{34} \right)
  \end{align*}
  We can convert this constraint into linear inequalities using the recipies we have seen previously in this survey. Adding the inequality into the ILP from above will give us the tree in Figure~\ref{fig:flow-constraints-example}(c).

  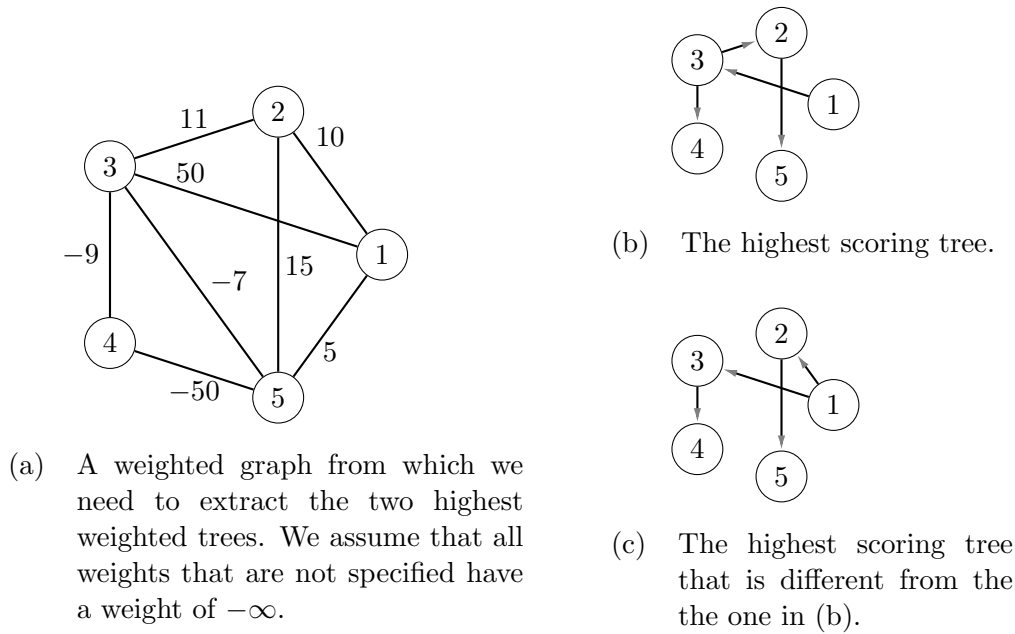
\begin{figure}
    \centering
    \begin{tikzpicture}

      \foreach \i in {1,...,5} {
        \pgfmathsetmacro\r{(\i-1)*(360/5)}
        \node[draw,circle] (n\i) at (\r:2) {$\i$};
      }

      \draw[thick] (n1) -- node[midway, above=10pt] {$10$} (n2);
      \draw[thick] (n1) -- node[near end, above] {$50$} (n3);    
      \draw[thick] (n1) -- node[midway, below=2pt] {$5$} (n5);
      \draw[thick] (n2) -- node[midway, above] {$11$} (n3);
      \draw[thick] (n2) -- node[midway, yshift=-4pt, xshift=8pt] {$15$} (n5);
      \draw[thick] (n3) -- node[midway, left] {$-9$} (n4);
      \draw[thick] (n3) -- node[near end, above=10pt] {$-7$} (n5);
      \draw[thick] (n4) -- node[midway, below] {$-50$} (n5);            

      \node[xshift=-0.5cm, below=0.2cm of n5] {
        \begin{minipage}{0.4\linewidth}
          \begin{tabular}{lp{0.9\linewidth}}
            (a) & A weighted graph from which we need to extract the two highest weighted trees. We assume that all weights that are not specified have a weight of $-\infty$.
          \end{tabular}
        \end{minipage}
      };

      \foreach \i in {1,...,5} {
        \pgfmathsetmacro\r{(\i-1)*(360/5)}
        \node[draw,circle] (t\i) at ([xshift=7cm, yshift=2cm]\r:1) {$\i$};
      }

      \draw[thick, -{Latex[length=2mm,width=1mm,gray]}] (t1) --  (t3);
      \draw[thick, -{Latex[length=2mm,width=1mm,gray]}] (t3) --  (t2);
      \draw[thick, -{Latex[length=2mm,width=1mm,gray]}] (t2) --  (t5);
      \draw[thick, -{Latex[length=2mm,width=1mm,gray]}] (t3) --  (t4);

      \node[below=0.2cm of t5] {
        \begin{minipage}{0.3\linewidth}
          \begin{tabular}{lp{0.9\linewidth}}
            (b) & The highest scoring tree.
          \end{tabular}
        \end{minipage}};

      \foreach \i in {1,...,5} {
        \pgfmathsetmacro\r{(\i-1)*(360/5)}
        \node[draw,circle] (tt\i) at ([xshift=7cm,yshift=-2cm]\r:1) {$\i$};
      }

      \draw[thick, -{Latex[length=2mm,width=1mm,gray]}] (tt1) --  (tt2);
      \draw[thick, -{Latex[length=2mm,width=1mm,gray]}] (tt1) --  (tt3);
      \draw[thick, -{Latex[length=2mm,width=1mm,gray]}] (tt2) --  (tt5);
      \draw[thick, -{Latex[length=2mm,width=1mm,gray]}] (tt3) --  (tt4);

      \node[below=0.2cm of tt5] {
        \begin{minipage}{0.3\linewidth}
          \begin{tabular}{cp{0.9\linewidth}}
          (c) & The highest scoring tree that is different from the the one in (b).
          \end{tabular}
        \end{minipage}
      };

    \end{tikzpicture}
    \caption{An undirected graph to illustrate the spanning tree constraints. The goal is to find the two highest scoring trees spanning the nodes in subfigure (a). Example~\ref{eg:spanning-tree} shows how to generate the two trees in subfigures (b) and (c) incrementally. The directed edges in the tree show the direction of commodity flow in the solutions to the mixed integer programs.}
    \label{fig:flow-constraints-example}
  \end{figure}

\end{example}


\subsection{Graph Connectivity}
\label{sec:graph-connectivity}

Our second complex building block involves distilling a \emph{connected}
sub-graph from a given graph. Suppose our graph at hand is directed and we seek
to select a sub-graph that spans all the nodes and is connected.  We can reduce
this to the spanning tree constraint by observing that any connected graph
should contain a spanning tree. This observation gives us the following solution
strategy: Construct an auxiliary problem (i.e, finding a spanning tree) whose
solution will ensure the connectivity constraints we need. 

Let inference variables $\var{ij}$ denote the decision that the edge $(i,j)$ is
selected. To enforce the connectivity constraints, we will introduce auxiliary
Boolean inference variables $z_{ij}$ (with zero objective coefficients) for
every edge $(i,j)$ or $(j,i)$ that is in the original graph. In other words, the auxiliary variables we introduce are undirected.

Using these auxiliary variables, we can state the connectivity requirement as follows:
\begin{enumerate}
\item The inference variables $z_{ij}$ form a spanning tree over the nodes.
\item If $z_{ij}$ is \true, then either the edge $(i,j)$ or the edge $(j,i)$
  should get selected.
\end{enumerate}

We can write these two requirements using the building blocks we have already
seen.
\begin{constraint}{Find a connected spanning sub-graph of the nodes $1,2,\cdots,n$}
  \mbox{{\sc Spanning Tree} constraints over variables } z_{ij}, \\
  \mbox{for every } (i,j) \mbox{ such that } i < j, \quad\quad z_{ij} \rightarrow \var{ij}  \vee \var{ji}.
\end{constraint}
Each of these constraints can be reduced to a collection of linear inequalities
using the tools we have seen so far. We will see an example of how a variant of
this recipe can be used in \S\ref{sec:events-relations}. In the construction above, the $z$'s help set up the auxiliary spanning tree problem. Their optimal values are typically disregarded, and it is the assignment to the $\var{}$'s that constitute the solution to the original problem.


\subsection{Other Graph Problems}
\label{sec:other-graph-problems}

In general, if the problem at hand can be written as a known and
tractable graph problem, then there are various efficient ways to
instantiate linear inequalities that encode the structure of the
output graph. We refer the reader to resources such as
\citet{papadimitriou1982combinatorial}, \citet{magnanti1995optimal} and \citet{schrijver1998theory} for further
reference. We also refer the reader to the AD\textsuperscript{3}
algorithm~\citep{martins2015ad3} that supports the coarse decomposition of
inference problems to take advantage of graph algorithms directly.


\subsection{Soft Constraints}
\label{sec:soft-constraints}
The constraints discussed so far in this survey are hard
constraints. That is, they prohibit certain assignments of the
decision variables. In contrast, a \emph{soft constraint} merely
penalizes assignments that violates them rather than disallowing
them.
Soft constraints can be integrated into the integer linear programming
framework in a methodical fashion. \citet{srikumar2013soft} explains
the process of adding soft constraints into ILP inference. Here we
will see a brief summary. 

As before, suppose we have an inference problem expressed as an integer linear
program:
\begin{eqnarray*}
  \max_{\vars} & \Sum_i \coeff{i}\var{i}  \\
  \mbox{s.t. } & \vars \in \sY,         \\
               & \var{i} \in \{0, 1\}.
\end{eqnarray*}
Here, the requirement that $\vars \in \sY$ is assumed to be stated as
linear inequalities. However, as we have seen in the previous
sections, they could be equivalently stated as Boolean expressions. 

If, in addition to the existing constraint, we have an additional
Boolean constraint $C(\by)$ written in terms of inference variables
$\by$. Instead of treating this as a hard constraint, we only wish to
penalize assignments $\by$ that violate this constraint by a penalty
term $\rho_C$. We will consider the case where $\rho_C$ is independent
of $\by$. To address inference in such a scenario, we can introduce a
new Boolean variable $z$ that tracks whether the constraint is not
satisfied. That is,
\begin{equation}
  \label{eq:soft-constraint-variable}
  z \leftrightarrow \neg C(\by).
\end{equation}

If the constraint is not satisfied, then the corresponding assignment
to the decision variables should be penalized by $\rho_C$. We can do
so by adding a term $-z\rho_C$ to the objective of the original
ILP. Since the constraint \eqref{eq:soft-constraint-variable} that
defines the new variable $z$ is also a Boolean expression, it can be
converted into a set of linear inequalities.

This procedure gives us the following new ILP that incorporates the
soft constraint:
\begin{eqnarray*}
  \max_{\vars, z} & \Sum_i \coeff{i}\var{i}  - z\rho_C                                    \\
  \mbox{s.t. }    & \vars \in \sY,                                                        \\
                  & z \leftrightarrow \neg C(\by),                                        \\
                  & \var{i}, z \in \{0, 1\}.
\end{eqnarray*}
We can summarize the recipe for converting soft constraints into
larger ILPs below:
\begin{constraint}{Soft constraint $C(\by)$ with a penalty $\rho_C$}
                  & \mbox{Add a Boolean variable $z$ to the objective with coefficient $-\rho_C$} \\
                  & \mbox{Add constraint } z \leftrightarrow \neg C(\by)
\end{constraint}



\section{Worked Examples}
\label{sec:worked-examples}

In this section, we will work through two example NLP tasks that use
the framework that we have seen thus far. First, we will look at the
problem of predicting sequences, where efficient inference
algorithms exist. Then, we will see the task of predicting
relationships between events in text, where we need the full ILP
framework even for a simple setting.

\subsection{Sequence Labeling}
\label{sec:sequence-labeling}
Our first example is the problem of sequence labeling. Using the tools
we have seen so far, we will write down prediction in a first order
sequence model as an integer linear program.

\begin{example}[Sequence Labeling]
  Suppose we have a collection of $n$ categorical decisions, each of
  which can take one of three values
  $\sL = \{\lbl{a}, \lbl{b}, \lbl{c}\}$. We can think of these $n$
  decisions as slots that are waiting to be assigned one the three
  labels. Each slot has an intrinsic preference for one of the three
  labels. Additionally, the label at each slot is influenced by the
  label of the previous slot. The goal of inference is to find a
  sequence of labels that best accommodates both the intrinsic
  preferences of each slot and the influence of the neighbors.

  Let us formalize this problem. There are two kinds of scoring
  functions. The decision at the $i^{th}$ slot is filled with a label
  $\lbl{L}$ is associated with an {\em emission} score
  $\coefflab{i}{\lbl{L}}$ that indicates the intrinsic preference of
  the slot getting the label. Additionally, pairs of decisions in the
  sequence are scored using {\em transition scores}. That is, the
  outcome that the $i^{th}$ label is $\lbl{L_1}$ and the $j^{th}$
  label is $\lbl{L_2}$ is jointly scored using
  $\coeff{\lbl{L_1},\lbl{L_2}}$. (Notice that the transition
  score is independent of $i$ in this formulation.) Now, our goal is
  find a label assignment to all $n$ slots that achieves the maximum
  total score.

  \begin{figure}
    \centering
    \begin{tikzpicture}
      \pgfmathtruncatemacro{\N}{5}

      \foreach \i in {1,...,\N}
      {
        \pgfmathtruncatemacro{\x}{\i * 2 - 1}

        \node[draw, circle] (y\i) at (\x, 2) {$\var{\i}$};

        \node[draw, rectangle, fill=black] (ce\i) at (\x, 1) {};

        \node[below =0.05cm of ce\i] {{\footnotesize $\coeff{\i}$}};
        
        \draw (y\i) -- (ce\i);
      }

      \foreach \i in {2,...,\N}
      {
        \pgfmathtruncatemacro{\prev}{\i - 1}
        
        \pgfmathtruncatemacro{\x}{2 * ( \i - 1)}

        \node[draw,rectangle, fill=black] (ct\prev\i) at (\x, 2) {};

        \node[above =0.05cm of ct\prev\i] {{\footnotesize $\coeff{\prev,\i}$}};

        \draw (y\prev) -- (ct\prev\i) -- (y\i);
      }

    \end{tikzpicture}
    \caption{An example factor graph for a sequence model. This figure
      illustrates the case of five decisions in a sequence.  Circles
      denote random variables whose assignment we seek and the squares
      represent factors or scoring functions as described in the
      text.}
    \label{fig:seq}
  \end{figure}
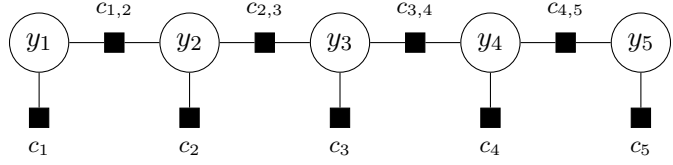

  Figure~\ref{fig:seq} gives the usual pictorial representation of
  this predictive problem.  A first-order sequence labeling problem of
  this form is ubiquitous across NLP for tasks such as part-of-speech
  tagging, text chunking and various information extraction problems.
  There are different ways to frame this problem as an ILP. We will
  employ one that best illustrates the use of the techniques we have
  developed so far.

  First, let us start with the decision variables. There are two kinds
  of decisions---emissions and transitions---that contribute to the
  total score. Let $\varlab{i}{\lbl{L}}$, scored by
  $\coefflab{i}{\lbl{L}}$, denote the decision that the $i^{th}$ label
  is $\lbl{L}$. Let $\varlab{i}{\lbl{L_1},\lbl{L_2}}$ denote
  the decision that the $i^{th}$ label is $\lbl{L_1}$ and the next one
  is $\lbl{L_2}$.  This transition is scored by
  $\coeff{\lbl{L_1},\lbl{L_2}}$. These variables and their
  associated scores give us the following objective function for the
  inference:
  \begin{equation}
    \label{eq:seq-objective}
    \max_{\vars} \Sum_{i=1}^n \Sum_{\lbl{L} \in \sL}\coefflab{i}{\lbl{L}} \cdot \varlab{i}{\lbl{L}}  + %
    \Sum_{i=1}^{n-1} \Sum_{\lbl{L_1},\lbl{L_2} \in \sL} \coeff{\lbl{L_1},\lbl{L_2}} \cdot \varlab{i}{\lbl{L_1},\lbl{L_2}}.
  \end{equation}
  Note that the objective simply accumulates scores from {\em every}
  possible decision that can be made during inference. For the sake of
  simplicity, we are ignoring initial states in this discussion, but
  they can be easily folded into the objective.
  
  Now that the inference variables are defined, we need to constrain
  them. We have two kinds of constraints:

  \begin{enumerate}
  \item Each slot can take exactly one label in
    $\sL = \{\lbl{a}, \lbl{b}, \lbl{c}\}$. Once again, we instantiate
    the {\sc Multiclass Classification as an ILP} construction (\S
    \ref{sec:disjunctions}) to get
    \begin{equation}
      \label{eq:seq-unique-labels}
      \forall i \in \{1, 2, \cdots, n\}; \Sum_{L \in \sL}\varlab{i}{\lbl{L}} = 1.
    \end{equation}
    These equations give us $n$ linear constraints in all.
  \item The transition decisions and the emission decisions should
    agree with each other. Written down in logic, this condition can
    be stated as:
    \begin{align*}
       \forall~i\in \{1, 2, \cdots, n\};                                                                  
      \quad\forall~\lbl{L_1},\lbl{L_2} \in \sL;                                                               
       \quad\varlab{i}{\lbl{L_1},\lbl{L_2}} \leftrightarrow \varlab{i}{\lbl{L_1}} \wedge \varlab{i+1}{\lbl{L_2}}  
    \end{align*}
    Together, these $n|\sL|^2$ constraints ensure that the output is a
    valid sequence. Since each of them is a conjunctive biconditional
    form (\S \ref{sec:complex-implications}), we get the following
    linear inequalities representing the constraints:
    \begin{eqnarray}
      \forall i, \lbl{L_1}, \lbl{L_2}; & -2\varlab{i}{\lbl{L_1},\lbl{L_2}}+\varlab{i}{\lbl{L_1}} + \varlab{i+1}{\lbl{L_2}}\geq 0 \label{eq:seq-consistency-1}\\
                                       & \varlab{i}{\lbl{L_1},\lbl{L_2}}-\varlab{i}{\lbl{L_1}} - \varlab{i+1}{\lbl{L_2}}\geq -1  \label{eq:seq-consistency-2}
    \end{eqnarray}
    In all, we get $2n|\sL|^2$ linear inequalities to represent these
    consistency constraints.
  \end{enumerate}

  The objective~\eqref{eq:seq-objective} and the
  constraints~\eqref{eq:seq-unique-labels},~\eqref{eq:seq-consistency-1}
  and~\eqref{eq:seq-consistency-2} together form the integer linear program for
  sequence labeling.
  
  It is important to note once again that here, we are only using the integer
  linear programs as a declarative language to state inference problems, not
  necessarily for solving them.  Specifically for the sequence labeling problem
  framed as a first-order Markov model, the Viterbi algorithm offers a
  computationally efficient solution to the inference problem. However, we may
  wish to enforce constraints that renders the Viterbi algorithm unusable.
  
  The strength of the ILP formulation comes from the flexibility it
  gives us. For example, consider the well-studied problem of
  part-of-speech tagging.  Suppose, we wanted to only consider
  sequences where there is at least one $\lbl{verb}$ in the final
  output. It is easy to state this using the following constraint:
  \begin{equation}
    \label{eq:seq-extra-constraint}
    \Sum_{i=1}^n \varlab{i}{\lbl{verb}} \geq 1.
  \end{equation}
  With this constraint, we can no longer use the vanilla Viterbi algorithm for
  inference. But, by separating the declaration of the problem from the
  computational strategies for solving them, we can at least write down the
  problem formally, perhaps allowing us to use a different algorithm, say
  Lagrangian
  relaxation~\citep{everettiii1963generalized,lemarechal2001lagrangian}, or a
  call to a black box ILP solver for solving the new inference problem.
\end{example}


\subsection{Recognizing Event-Event Relations}
\label{sec:events-relations}
Our second example involves identifying relationships between events in
text. While the example below is not grounded directly in any specific
instantiation of the task, it represents a simplified version of the inference
problem addressed by~\citet{berant2014modeling,NFWR18,NWPR18,wang2020joint}.

\begin{example}[Event-Event Relations]
  Suppose we have a collection of events denoted by
  $E = \{e_1, e_2,\cdots,e_n\}$ that are attested in some text. Our
  goal is to identify causal relationships between these events.
  That is, for any pair of events $e_i$ and $e_j$, we seek a
  directed edge that can be labeled with one of a set of labels
  $R =$ \{{\sc Cause}, {\sc Prevent}, {\sc None}\} respectively
  indicating that the event $e_i$ causes, prevents or is unrelated
  to event $e_j$.

  For every pair of events $e_i$ and $e_j$, we will introduce decision
  variables $\varlab{ij}{r}$ for each relation $r \in R$ denoting that
  the edge $(i,j)$ is labeled with the relation $r$. Each decision may
  be assigned a score $\coefflab{ij}{r}$ by a learned scoring
  function. Thus, the goal of inference is to find a score maximizing
  set of assignments to these variables. This gives us the following
  objective:
  \begin{equation}
    \label{eq:events-relations:obj}
    \Sum_{e_i, e_j \in E} \Sum_{r \in R} \coefflab{ij}{r}\cdot\varlab{ij}{r}. 
  \end{equation}
  Suppose we have three sets of constraints that restrict the set of
  possible assignments to the inference variables. These constraints
  are a subset of the constraints used to describe biological
  processes by~\citet{berant2014modeling}.
  
  \newcommand{\None}{\mbox{\footnotesize \sc None}}
  \newcommand{\Prevent}{\mbox{\footnotesize \sc Prevent}}
  \newcommand{\Cause}{\mbox{\footnotesize \sc Cause}}

  \begin{enumerate}
  \item Each edge should be assigned exactly one label in $R$. This is
    the {\sc Multiclass Classification as an ILP} construction, giving
    us
    \begin{equation}
      \label{eq:events-relations:unique-labels}
      \forall e_i, e_j \in E, \Sum_{r \in R} \varlab{ij}{r} = 1.
    \end{equation}
  \item If an event $e_i$ causes or prevents $e_j$, then $e_j$ can
    neither cause nor prevent $e_i$. In other words, if a {\sc Cause}
    or a {\sc Prevent} relation is selected for the $(i,j)$ edge, then
    the {\sc None} relation should be chosen for the $(j,i)$ edge. We
    can write this as a logical expression as:
    \begin{equation*}
      \forall e_i, e_j \in E, \varlab{ij}{\Cause} \vee \varlab{ij}{\Prevent} \rightarrow \varlab{ji}{{\None}}.
    \end{equation*}
    This is an example of a disjunctive implication (\S
    \ref{sec:complex-implications}), which we can write using linear
    inequalities as:
    \begin{equation}
      \label{eq:event-relations:causality}
      \forall e_i, e_j \in E, -\varlab{ij}{\Cause} - \varlab{ij}{\Prevent} +  2\varlab{ji}{\None} \geq 0.
    \end{equation}
  \item The events should form a connected component using the non-{\sc None}
    edges. This constraint invokes the graph connectivity construction from \S
    \ref{sec:graph-connectivity}.  To instantiate the construction, let us
    introduce auxiliary Boolean variables $z_{ij}$ that indicates that the
    events $e_i$ and $e_j$ are connected with an edge that is not labeled {\sc
      None} in at least one direction, i.e., the edge from $e_i$ to $e_j$ or the one in the other direction has a non-\None label. As before, let $\flow{ij}$ denote the
    non-negative real valued flow variables along a directed edge
    $(i,j)$. Following \S \ref{sec:graph-connectivity}, we will require that the
    $z_{ij}$'s form a spanning tree.
    
    First, the auxiliary variables $z_{ij}$ should correspond to events $e_i$
    and $e_j$ that are connected by a non-{\sc None} edge in either
    direction. That is,
    \begin{align}      
      \forall e_i, e_j \in E~\text{where}~i <j,~ & z_{ij} \rightarrow \p{\exists~ r \neq \None,~\text{s.t.}~\varlab{ij}{r} \vee \varlab{ji}{r}}, \label{eq:event-relations:auxiliary:1} 
    \end{align}
    The existential form on the right hand side of the implication can
    be written as a disjunction, thus giving us a disjunctive
    implication. For brevity, we will not expand these Boolean
    expressions into linear inequalities.

    Second, an arbitrarily chosen event $e_1$ sends out $n-1$ units of
    flow, and each event consumes one one unit of flow.
    \begin{eqnarray}
                         & \sum_j \flow{1j} - \sum_j\flow{j1} = n- 1, \label{eq:events-relations:flow-start} \\
      \forall e_i \in E, & \sum_j \flow{ij} - \sum_j\flow{ji} = 1. \label{eq:events-relations:flow-balance}
    \end{eqnarray}

    Third, the commodity flow should only happen along the edges that
    are selected by the auxiliary variables.
    \begin{equation}
      \label{eq:events-relations:flow-aux-connection}
      \forall e_i, e_j \in E~\text{where}~i <j,~\flow{ij} \leq (n-1) z_{ij}
    \end{equation}

    Finally, the auxiliary variables should form a tree. That is,
    exactly $n-1$ of them should be selected.
    \begin{equation}
      \label{eq:events-relations:tree}
      \Sum_{i,j} z_{ij} = n-1.
    \end{equation}
  \end{enumerate}

  We can write the final inference problem as the problem of
  maximizing the objective \eqref{eq:events-relations:obj} with
  respect to the inference variables $\vars$, the auxiliary variables
  $z_{ij}$ and the flow variables $\flow{ij}$ subject to the
  constraints listed in
  \Crefrange{eq:events-relations:unique-labels}{eq:events-relations:tree}.
  Of course, the decision variables $\vars$ and the auxiliary
  variables $z_{ij}$ are $0$-$1$ variables, while the flow variables
  are non-negative real valued ones.
\end{example}


\section{Final Words}
\label{sec:final-words}

We have seen a collection of recipes that can help to encode inference
problems as instances of integer linear programs. Each recipe focuses
on converting a specific kind of predicate  into one or more linear
inequalities that constitute the constraints for the discrete optimization
problem.
The conversion of predicates to linear inequalities is deterministic and, in
fact, can be seen as a compilation step, where the user merely specifies
constraints in first-order logic and an inference compiler produces efficient
ILP formulations. Some programs that allow declarative specification of
inference include Learning Based Java~\citep{rizzolo2012learning},
Saul~\cite{kordjamshidi2016better} and DRaiL~\cite{pacheco2021modeling}.

It should be clear from this tutorial-style survey that there may be
multiple ways to encode the same inference problem as integer
programs. The best encoding may depend on how the integer program is
solved. Current solvers (circa 2022) seem to favor integer programs with
fewer constraints that are dense in terms of the number of variables
each one involves. To this end, we saw two strategies: We either
collapsed multiple logical constraints that lead to sparse
inequalities to fewer dense ones, or formulated the problem in terms
of known graph problems.

While it is easy to write down inference problems, it is important to
keep the computational properties of the inference problem in
mind. The simplicity of design can make it easy to end up with large
and intractable inference problems. For example, for the event
relations example from \S \ref{sec:events-relations}, if we had tried
to identify both the events and their relations using a single integer
program (by additionally specifying event decision variables), the
approach suggested here can lead to ILP instances that are difficult
to solve with current solvers.

A survey on using integer programming for modeling inference would
be remiss without mentioning techniques for solving the integer
programs. The easiest approach is to use an off-the-shelf
solver. Currently, the fastest ILP solver is the Gurobi
solver;\footnote{http://www.gurobi.com} other solvers include the
CPLEX
Optimizer,\footnote{https://www.ibm.com/products/ilog-cplex-optimization-studio}
the FICO
Xpress-Optimizer,\footnote{http://www.fico.com/en/products/fico-xpress-optimization-suite}
{\tt lp\_solve},\footnote{https://sourceforge.net/projects/lpsolve}
and GLPK.\footnote{https://www.gnu.org/software/glpk/} The advantage
of using off-the-shelf solvers is that we can focus on the problem
at hand. However, using such solvers prevents us from using
task-driven specialized strategies for inference, if they
exist. Sometimes, even though we can write the inference problem as
an ILP, we may be able to design an efficient algorithm for solving
it by taking advantage of the structure of the problem.
Alternatively, we can relax the problem by simply dropping the
$\{0,1\}$ constraints over the inference variables and instead
restricting them to be real valued in the range $[0,1]$. We could also
employ more sophisticated relaxation methods such as Lagrangian
relaxation~\citep{everettiii1963generalized,lemarechal2001lagrangian,geoffrion2010lagrangian,chang2011exact},
dual decomposition~\citep{rush2012tutorial,rush2010dual,koo2010dual},
or the augmented Lagrangian
method~\citep{martins2011augmented,martins2011dual,meshi2011alternating,martins2015ad3}. 

The ability to write down prediction problems in a declarative fashion (using
predicate logic or equivalently as ILPs) has several advantages. First, we can
focus on the definition of the task we want to solve rather than the algorithmic
details of how to solve it. Second, because we have a unifying language for
reasoning about disparate kinds of tasks, we can start reasoning about
properties of inference in a task-independent fashion. For example, using such
an abstraction, we can amortize inference costs over the lifetime of the
predictor~\citep{srikumar2012amortizing,kundu2013marginbased,chang2015structural,pan2018learning}.

Finally, recent successes in NLP have used neural models with pre-trained
representations such as BERT~\citep{devlin2019bert},
RoBERTa~\citep{liu2019roberta} and others. The unification of such neural
networks and declarative modeling with logical constraints is an active area of
research today~\citep[inter
alia]{xu2018semantic,li2019augmenting,li2019logicdriven,fischer2019dl2,nandwani2019primal,li2020structured,wang2020joint,asai2020logicguided,giunchiglia2021multilabel,grespan2021evaluating,pacheco2021modeling,ahmed2022pylon}. This
area is intimately connected with the area of neuro-symbolic modeling which
seeks to connect neural models with symbolic reasoning. We refer the reader
to~\citet{garcez2020neurosymbolic,kautz2022third,pacheco2022ns4nlp} for recent
perspectives on the topic. The declarative modeling strategy supported by the
kind of inference outlined in this tutorial may drive the integration of complex
symbolic reasoning with expressive neural models, which poses difficulties for
current state-of-the-art models.


\bibliography{cited,ccg}

\bibliographystyle{plainnat}

\end{document}